\documentclass{article}

\usepackage{microtype}
\usepackage{graphicx}
\usepackage{subfigure}
\usepackage{booktabs} 
\usepackage[utf8]{inputenc} 
\usepackage[T1]{fontenc}    
\usepackage{hyperref}       
\usepackage{url}            
\usepackage{booktabs}       
\usepackage{amsfonts}       
\usepackage{nicefrac}       
\usepackage{microtype}      
\usepackage{graphicx}
\usepackage{amsmath}
\usepackage{float}
\usepackage{amsmath}
\usepackage{optidef}
\usepackage{hhline}
\usepackage{multirow}
\usepackage{enumitem}
\usepackage{xcolor}
\usepackage{wrapfig}
\usepackage{caption}

\usepackage{hyperref}

\usepackage[accepted]{icml2019}

\icmltitlerunning{Machine learning for AC Optimal Power Flow}

\begin{document}

\twocolumn[
\icmltitle{Machine Learning for AC Optimal Power Flow}




\begin{icmlauthorlist}
\icmlauthor{Neel Guha}{cmu}
\icmlauthor{Zhecheng Wang}{stanford}
\icmlauthor{Matt Wytock}{gm}
\icmlauthor{Arun Majumdar}{stanford}
\end{icmlauthorlist}

\icmlaffiliation{cmu}{Carnegie Mellon University}
\icmlaffiliation{stanford}{Stanford University}
\icmlaffiliation{gm}{Gridmatic, Inc.}
\icmlcorrespondingauthor{Neel Guha}{neelguha@gmail.com}

\icmlkeywords{Machine Learning, ICML}
\vskip 0.3in
]
\printAffiliationsAndNotice{} 

\begin{abstract}
We explore machine learning methods for AC Optimal Powerflow (ACOPF) - the task of optimizing power generation in a transmission network according while respecting physical and engineering constraints. We present two formulations of ACOPF as a machine learning problem: 1) an \textit{end-to-end prediction} task where we directly predict the optimal generator settings, and 2) a \textit{constraint prediction} task where we predict the set of active constraints in the optimal solution. We validate these approaches on two benchmark grids.  
\end{abstract}

\section{Introduction}

The Optimal Power Flow problem (OPF) consists of determining the optimal operating levels for different generators within a transmission network in order to meet the demand that is changing over space and time. An established area of research in both power systems and operations, OPF is applied every day in the management and regulation of power grids around the world.  In this work, we hope to obtain real-time approximate solutions to the OPF problem using machine learning.

The classical formulation of ACOPF (presented in Section \ref{section:formal}) is a challenging non-convex and NP-hard problem \cite{bienstock2015strong}. In addition to minimizing generator costs, solutions must adhere to physical laws governing power flow (i.e. Kirchhoff's voltage law) and respect the engineering limits of the grid. As a result, ACOPF is computationally intractable under the demands of daily grid management. In order to account for rapid fluctuations in power demand and supply, grid operators must solve ACOPF over the entire grid (comprising of tens of thousands of nodes) every five minutes \footnote{The addition of renewable sources of energy (wind, solar, etc)  adds more unpredictability and is a motivation for improved techniques for ACOPF} \cite{cain2012history}.  Most traditional approaches (genetic algorithms, convex relaxations, etc) either fail to converge within this time frame or produce suboptimal solutions. In order to practically manage the grid, operators solve a linearized version of ACOPF practice known as DC Optimal Power Flow (DCOPF).  However, DCOPF presents a number of issues. True grid conditions can deviate from the linear assumptions imposed by DCOPF, increasing the likelihood of instability and grid failure \cite{frank2016introduction}. Relying on DCOPF also has significant implications for climate change. A 2012 report from the Federal Energy Regulatory Commission estimated that the inefficiencies induced by approximate-solution techniques may cost billions of dollars and release unnecessary emissions \cite{cain2012history}. Determining an efficent solution for ACOPF could also be adapted  to \textit{combined economic emission dispatch} (CEED) - a variant of OPF which incorporates a per-generator emissions cost into the classic objective function \cite{venkatesh2003comparison}.

In this paper, we observe that it should be possible to learn a model that can predict an accurate solution over a fixed grid topology/constraint set. Intuitively, we expect some measure of consistency in the solution space - similar load distributions should correspond to similar generator settings. This suggests an underlying structure to the ACOPF problem, which a machine learning model can exploit.  

Machine learning present several advantages. Neural networks have demonstrated the ability to model extremely complicated non-convex functions, making them highly attractive for this setting. A model could be trained off-line on historic data and used in real-time to make predictions on an optimal power setting. In this work, we explore two applications of machine learning for OPF:

\textbf{1. End-to-end}: Train a model to directly predict the optimal generator setting for a given load distribution. This is challenging, as the model's output must be adherence with physical laws/engineering limits.\\

\textbf{2. Constraint prediction}: Train a model to predict which constraints are active (i.e at equality) in the optimal solution. Knowing this active set can be used to warm start existing approaches (i.e. interior point methods) and reduce solution time.

\section{Related Work}
Prior work has explored different applications of machine learning on the grid. This includes work on estimating active constraints for DCOPF \cite{ng2018statistical, misra2018learning}, predicting grid failures \cite{rudin2012machine}, or choosing between traditional solvers \cite{king2015network}. Machine learning has also been applied to related variants of the OPF problem, including automated grid protection \cite{donnot2017introducing}, price proxy prediction \cite{DBLP:journals/corr/CanyasseDM16}, or private information recovery \cite{dontiinverse}. To the extent of our knowledge, there has been limited work on direct applications of deep learning towards ACOPF.

\section{Method} \label{section:formal}
We now present the traditional ACOPF problem, and describe how to formalize it as a machine learning task\cite{frank2012primer}. For a fixed grid topology $\mathcal{G}$, let $\mathbf{N}$ denote the set of buses (nodes), $\mathbf{L}$ denote the set of branches (edges), and $\mathbf{G} \subseteq \mathbf{N}$ denote the set of controllable generators. For bus $i$, we enumerate $P_i^G$ (real power injection), $Q_i^G$ (reactive power injection), $P_i^L$ (real power demand), $Q_i^L$ (reactive power demand), $V_i$ (voltage magnitude), and $\delta_i$ (voltage angle).
the power demand at AC OPF can be framed as:

\begin{mini!}|l|[3]
    {P_i^G}{\sum_{i \in \mathbf{G}}C_i(P_i^G)}
    {}{}
    \addConstraint{P_i(V, \delta)}{= P_i^G - P_i^L,\quad}{\forall i \in \mathbf{N}}
    \addConstraint{Q_i(V, \delta)}{= Q_i^G - Q_i^L,\quad}{\forall i \in \mathbf{N}}
    \addConstraint{P_i^{G,\min} \leq P_i^G \leq P_i^{G,\max}}{,\quad}{\forall i \in \mathbf{G}}
    \addConstraint{Q_i^{G,\min} \leq Q_i^G \leq Q_i^{G,\max}}{,\quad}{\forall i \in \mathbf{G}}
    \addConstraint{V_i^{\min} \leq V_i \leq V_i^{\max}}{,\quad}{\forall i \in \mathbf{N}}
    \addConstraint{\delta_i^{\min} \leq \delta_i \leq \delta_i^{\max}}{,\quad}{\forall i \in \mathbf{N}}
\end{mini!}

Where (1a) typically represents a polynomial cost function, (1b)-(1c) corresponds to the power flow equations, and (1d)-(1g) represent operational limits on real/reactive power injections, nodal voltage magnitude, and nodal voltage angles\footnote{A single reference bus ("slack" bus) is fixed to $\tilde{V} =1.0\angle0$} respectively. More recent settings of OPF - including ours-  also include limits on branch currents. These are outlined in more detail by \citet{frank2012primer}. We now present two formalizations of AC OPF as a machine learning problem. In our setting, we assume that $P_i^L$ and $Q_i^L$ (real and reactive demand) are known across all $N$ buses.

\subsection{End-to-end Prediction}\label{section:e2e}
In this setting, we pose the AC OPF problem as a regression task, where we predict the grid control variables ($P_i^G$ and $V_i^G$) from the grid demand ($P^L_i$ and $Q^L_i$). These fix a set of equations with equal number of unknowns, which can be solved to identify the remaining state values for the grid. Formally, given a dataset of $n$ solved grids with load distributions $\mathbf{X} = \{[P_0^L, ..,P_\mathbf{N}^L, Q_0^L,...,Q_\mathbf{N}^L]\}_{i=1}^n$ and corresponding optimal generator settings $\mathbf{Y} = \{[P_0^G, ..,P_\mathbf{G}^G, V_0^G,...,V_\mathbf{G}^L]\}_{i=1}^n$, our goal is to learn $f_\theta: \mathcal{X}\rightarrow \mathcal{Y}$ which minimizes the mean-squared error between the optimal generator settings $\mathbf{Y}$ and the predicted generator settings $\tilde{Y}$. Solving for the remaining state variables can be posed as a \textit{power flow} problem, and reduces to finding $V_i^L$, $Q_i^G$, and $\delta_i$ such that (1b)-(1g) are satisfied. 

The central challenge in this setting is ensuring that the neural network's solution respects physical laws and engineering limits. Though provable guarantees may be difficult to make, we experiment by incorporating soft penalties into our loss function that encourage predictions to fall within legal limits. These correspond to linear penalties that activate when when (1d) and (1f) are violated. In future work we hope to explore more sophisticated (and robust) techniques for enforcing legality.

\subsection{Optimal Constraint Prediction}
Given that neural networks may learn solutions that violate physical constraints, and are thus untrustworthy in practical settings, we explore \textit{optimal constraint prediction} as formulated by \citet{misra2018learning}. In this setting, our model is trained to predict the set of constraints that are active in the optimal solution for some load distribution. A constraint is active if the corresponding state/control variable is at the maximum or minimum allowed value. As \citet{misra2018learning} describe, knowing the active set of constraints can be used to warm start a more traditional optimization method, and reduce time to convergence.

Formally, for each grid we define a constraint vector $y \in \mathbb{R}^{2G + 2N}$ corresponding to an enumeration of constraints (1d)-(1e), where $y_i = 1$ if the $i$-the constraint is active in the optimal solution, and $y_i = 0$ otherwise. We learn $f_\theta$ which maps from the load distribution $[P_0^L, ..,P_\mathbf{N}^L, Q_0^L,...,Q_\mathbf{N}^L]$ to this constraint vector. This corresponds to a multi-label classification problem. 

Optimal constraint prediction presents several advantages over end-to-end prediction. 
\begin{enumerate}[leftmargin=*]
    \item \textbf{Solver Speedup}: From an optimization perspective, knowing the set of active constraints equates to warm-starting, and can significantly speed-up more traditional algorithms like interior point methods, active set methods, simplex methods, and others. Quantifying this speedup is the focus of ongoing work. 
    \item \textbf{Reliability}: This setting reduces the risk of  a neural network producing a solution which violates physical laws/engineering limits. Because the physical and engineering constraints are enforced by the solver, an incorrect prediction will at worst increase solution time or lead to a suboptimal solution. In the end-to-end setting described in Section \ref{section:e2e}, incorrect predictions could destabilize the grid. 
    \item \textbf{Task complexity}: Classifying the set of active constraints is significantly easier than predicting a set of real valued targets. 
\end{enumerate}

\section{Results}
We validated approaches for end-to-end prediction and constraint prediction on IEEE 30-bus \footnote{\url{https://electricgrids.engr.tamu.edu/electric-grid-test-cases/ieee-30-bus-system/}} and 118-bus test cases\footnote{\url{https://electricgrids.engr.tamu.edu/electric-grid-test-cases/ieee-118-bus-system/}}. These test cases include predetermined constraints.

\subsection{Dataset Generation}
The IEEE test cases include a pre-calculated load distribution (denoted as $x^*$. In order to construct a dataset for each case, we repeatedly sample candidate load distributions $x' \sim \text{Uniform}((1-\delta)\cdot x^*, (1+\delta)\cdot x^*)$, for some fixed $\delta$. We identify $y'$ by solving the OPF problem for $x'$ via Matpower \cite{zimmerman2011matpower}. In some cases, the solver fails to converge, suggesting that the sampled $x'$ has no solution given the grid constraints. In this case, we discard $x'$. 

We generated 95000 solved grids for case118 and 812888 solved grids for case30 with $\delta = 0.1$ (a $10\%$ perturbation to the IEEE base demand). Interestingly, we observe that while 100\% of the samples generated for case118 were successfully solved, only $81.2\%$ of the samples for case30 were successfully solved. For all prediction tasks, we used a 90/10 train-test split and report results on the test set.

\subsection{End to end prediction}
We evaluate task performance along two metrics: 
\begin{itemize}[leftmargin=*]
    \item \textbf{Legality Rate}: The proportion of predicted grids which satisfy all engineering and physical constraints. 
    \item \textbf{Avg. Cost Deviation}: The average fractional difference between the cost of the predicted grid, and the cost of the true grid: $\dfrac{1}{n}\sum_i^n |1 - \dfrac{\text{pred cost}_i}{\text{true cost}_i}|$ over legal grids.
\end{itemize}
Roughly, this captures the reliability and optimality of a particular model. We examine a range of different architectures and training strategies. We performed a grid search considering models with 1-2 hidden layers, 128/256/512 hidden neurons, ReLU/Tanh activations. We also experimented with vanilla MSE loss, and a variant with linear penalties for constraint violations (described in Section \ref{section:e2e}). Each model was trained with Adam ($\text{lr} = 0.001$) until loss convergence, for a maximum of 2000 epochs.

\begin{table}[h!]
    \centering
    \renewcommand{\arraystretch}{0.5} 
    \begin{tabular}{lcc}
    \toprule
    \bf Grid & \bf Legality Rate & \bf Avg. Cost Deviation  \\ \midrule
    case30 & 0.51 & 0.002  \\ \midrule 
    case118 & 0.70  & 0.002  \\ \midrule
    \end{tabular}
    \captionsetup{justification=centering}
    \captionof{table}{End-to-end prediction performance. Average cost deviation is only reported for legal grids.}
    \label{table:e2e}
\end{table}

Table \ref{table:e2e} reports the best performance for each grid type. For case30, the optimal model was a two layer neural network with tanh activations, and no loss penalty. For case118, the optimal model was a three layer network with 512 hidden neurons, ReLU activations, and a constraint loss penalty. Interestingly, we observe better performance on case118 than case30. Though we would intuitively expect task difficulty to scale with grid size, this result suggests that other factors could affect a model's generalization ability. In particular, smaller grids could be less stable, and thus more likely to produce a wide range of (less predictable) behavior under varying demand distributions. We also observe that the cost of the optimal model predictions were within $1\%$ of the optimal cost.

\subsection{Constraint Prediction}
For constraint prediction, we evaluate performance in terms of accuracy (i.e. the proportion of constraints classified successfully). We perform a similar hyperparameter grid search and report the best results in Table \ref{table:cp}. 
\begin{table}[h!]
    \centering
    \renewcommand{\arraystretch}{0.5} 
    \begin{tabular}{lc}
    \toprule
    \bf Grid & \bf \% Accuracy \\ \midrule
    case30 & 0.99  \\ \midrule 
    case118 & 0.81    \\ \midrule
    \end{tabular}
    \captionsetup{justification=centering}
    \captionof{table}{Constraint prediction performance}
    \label{table:cp}
\end{table}

In general, we find neural networks to be highly successful at determining which the active constraint set. 

\section{Conclusion}
In this work, we presented two approaches that leverage machine learning for solving ACOPF. Preliminary experiments present promising results in both settings. In next steps, we hope to evaluate our methods on more complex grid architectures, and explore different approaches for incorporating grid constraints into our models.   


\bibliography{example_paper}
\bibliographystyle{icml2019}

\end{document}